\documentclass[10pt,conference]{IEEEtran}
\IEEEoverridecommandlockouts
\usepackage{amsmath,amssymb,amsfonts}
\usepackage{algorithmic}
\usepackage{graphicx}
\usepackage{textcomp}
\usepackage{xcolor}
\def\BibTeX{{\rm B\kern-.05em{\sc i\kern-.025em b}\kern-.08em
    T\kern-.1667em\lower.7ex\hbox{E}\kern-.125emX}}

\usepackage{graphicx}
\usepackage{tabularx}
\usepackage{xspace} 
\usepackage[many]{tcolorbox} 
\usepackage{ifthen}
\usepackage{amsmath,amssymb,amsfonts}
\usepackage{booktabs}
\usepackage{multirow}
\usepackage{graphicx}
\usepackage{colortbl}
\usepackage{listings}
\usepackage{soul}
\usepackage{flushend}
\usepackage{balance}
\usepackage{xcolor}
\usepackage[normalem]{ulem}
\usepackage[
    backend=bibtex,
    style=numeric-comp,
    sorting=nyt,
    giveninits=false,
    maxnames=3,
    natbib,
    defernumbers=true
]{biblatex}

\definecolor{lightgray}{rgb}{0.9, 0.9, 0.9}
\definecolor{lightblue}{rgb}{0.8, 0.9, 1.0}

\newboolean{showcomments}
\setboolean{showcomments}{true}
\ifthenelse{\boolean{showcomments}}{
    \newcommand{\nbc}[3]{
        {\colorbox{#3}{\bfseries\sffamily\scriptsize\textcolor{white}{#1}}}
        {\textcolor{#3}{\sf\small$\blacktriangleright$\textit{#2}$\blacktriangleleft$}}
    }
}{
    \newcommand{\nbc}[3]{}
}

\newif\ifrevision
\revisionfalse   

\ifrevision
  \newcommand{\rev}[1]{\textcolor{blue}{#1}}
  
  \NewEnviron{revision}{\textcolor{blue}{\BODY}}
  
  \newcommand{\del}[1]{\textcolor{red}{\sout{#1}}}

\else
  \newcommand{\rev}[1]{#1}
  \NewEnviron{revision}{\BODY}
  \newcommand{\del}[1]{}
\fi

\newcommand{\fontbetween}{\fontsize{6.9pt}{7.7pt}\selectfont}

\newcommand{\fnu}[1]{\footnote{\url{#1}}}
\newcommand{\quotes}[1]{``#1''}
\newcommand{\code}[1]{\texttt{#1}}

\newcommand{\acr}{AutoCodeRover\xspace}
\newcommand{\oh}{OpenHands\xspace}

\newcommand{\watson}{\textsf{Watson}\xspace}
\newcommand{\mmlu}{MMLU\xspace}

\newcommand{\llm}{LLM\xspace}
\newcommand{\llms}{LLMs\xspace}

\newcommand{\lrms}{LRMs\xspace}

\newcommand{\gptthreefiveturboinstruct}{GPT-3.5-turbo-instruct\xspace}
\newcommand{\gptthreefiveturbo}{GPT-3.5-turbo\xspace}
\newcommand{\gptfouromini}{GPT-4o-mini\xspace}
\newcommand{\swebenchlite}{SWE-bench-lite\xspace}
\newcommand{\swebench}{SWE-bench\xspace}

\newcommand{\promptexp}{PromptExp\xspace}
\newcommand{\django}{Django\xspace}

\newcommand{\fim}{FIM\xspace}
\newcommand{\repcot}{RepCoT\xspace}

\newcommand{\agentware}{Agentware\xspace}

\newtcolorbox{myframe}[1][]{
  enhanced,
  arc=0pt,
  outer arc=0pt,
  colback=white,
  boxrule=0.8pt,
  #1
}

\begin{document}
\title{Watson: A Cognitive Observability Framework for the Reasoning of LLM-Powered Agents}
\author{
    \IEEEauthorblockN{
        Benjamin Rombaut\IEEEauthorrefmark{1}, 
        Sogol Masoumzadeh\IEEEauthorrefmark{1}, 
        Kirill Vasilevski\IEEEauthorrefmark{1}, 
        Dayi Lin\IEEEauthorrefmark{1}, 
        Ahmed E. Hassan\IEEEauthorrefmark{3}
    }
    \IEEEauthorblockA{\IEEEauthorrefmark{1}Centre for Software Excellence, Huawei Canada}
    \IEEEauthorblockA{\IEEEauthorrefmark{3}Queen's University, Kingston, Canada}
}
\maketitle

\begin{abstract}
Large language models (\llms) are increasingly integrated into autonomous systems, giving rise to a new class of software known as \emph{\agentware}, where \llm-powered agents perform complex, open-ended tasks in domains such as software engineering, customer service, and data analysis. However, their high autonomy and opaque reasoning processes pose significant challenges for traditional software observability methods. To address this, we introduce the concept of \emph{cognitive observability}---the ability to recover and inspect the implicit reasoning behind agent decisions. We present \watson, a general-purpose framework for observing the reasoning processes of fast-thinking \llm agents without altering their behavior. \watson retroactively infers reasoning traces using prompt attribution techniques. We evaluate \watson in both manual debugging and automated correction scenarios across the \mmlu benchmark and the \acr \rev{and \oh} agent\rev{s} on the \swebenchlite dataset. In both static and dynamic settings, \watson surfaces actionable reasoning insights and supports targeted interventions, demonstrating its practical utility for improving transparency and reliability in \agentware systems.


\end{abstract}


\section{Introduction}
\label{sec:intro}

Agentic software powered by large language models (\llms)~\citep{hu2024automateddesignagenticsystems, change2024surveyevaluationllms, masterman2024landscapeemergingaiagent} (i.e., \emph{\agentware}~\citep{hassan_rethinking_2024}) is increasingly being adopted across a variety of domains such as autonomous software engineering~\citep{zhang_2024_autocoderover, devin_url}, customer support~\citep{haptik_url}, and data analytics~\citep{guo_2024_dsagent}. Unlike traditional software systems, \agentware operates with high autonomy, making decisions through opaque and implicit reasoning processes executed by the underlying \llms. While this enables greater flexibility and task generalization, it also introduces new challenges in observability, as developers can no longer rely on code-level traceability to understand or debug system behavior~\citep{hassan2024ainativesoftwareengineeringse, xu2024largelanguagemodelsreally}. As a result, ensuring the reliability, explainability, and controllability of such systems requires fundamentally new approaches to observing and interpreting their internal decision-making processes.

Traditional observability approaches such as logging, tracing, and performance monitoring are effective for deterministic software because developers can instrument code to trace execution and diagnose issues~\citep{li2022observability, sigelman2010dapper, data61_agentops}. However, these techniques fall short in the context of \agentware, where behavior is driven by non-deterministic prompt-based reasoning inside \llms. As these models tend to not expose their internal decision paths, developers are often left without insight into why an agent chose a particular action or produced a specific output. To address this gap, we propose a new concept called \emph{cognitive observability}, which \del{includes capturing the implicit reasoning processes of agents---not just what they do, but why they do it. This capability is essential for debugging, auditing, and improving \agentware, especially in dynamic, multi-agent, or safety-critical settings.}
\rev{extends observability beyond system performance to capture higher-level signals about an agent’s behavior and decision-making. Prior work has begun to explore aspects such as \emph{semantic feedback} and \emph{output integrity}, but the most underdeveloped dimension remains an agent’s \emph{reasoning trace}---the implicit steps it takes to reach a conclusion. Our work emphasizes this dimension, showing how surfacing reasoning traces provides critical support for debugging, auditing, and systematically improving \agentware in dynamic, multi-agent, and safety-critical settings.}

\del{
In this paper, we introduce the concept of \emph{cognitive observability} as a necessary extension to existing taxonomies of observability for \agentware~\citep{data61_agentops}, with a focus on understanding implicit agent reasoning. W}
\rev{Building on prior work that has examined semantic feedback and output integrity~\citep{data61_agentops}, we position our contribution within a broader taxonomy of \emph{cognitive observability}, which we define as encompassing these existing dimensions together with the underexplored notion of reasoning trace. Our emphasis is on this latter dimension: recovering the implicit steps an agent takes to reach its conclusions. To this end, w}e design and implement \watson, a general-purpose framework for observing the reasoning processes of fast-thinking LLM agents without modifying their behavior. We evaluate \watson in both manual debugging and automatic correction scenarios across the Massive Multitask Language Understanding (\mmlu)~\citep{hendrycks2021measuring} benchmark and the \acr~\citep{zhang_2024_autocoderover} \rev{and \oh~\citep{wang2025openhandsopenplatformai}} agent\rev{s} on the \swebenchlite~\citep{swebench_lite_url} benchmark. In both static and dynamic task settings, reasoning traces generated by \watson surface actionable insights and enable targeted interventions, demonstrating its utility both as a manual and an automated observability tool for \llm-powered systems.

The remainder of the paper is structured as follows: Section~\ref{sec:fm_observability} discusses observability for \agentware. 
Section~\ref{sec:watson} introduces \watson, our proposed framework. 
Section~\ref{sec:evaluation} provides a case study on applying \watson to agentic systems to demonstrate its practical applicability.
\rev{Section~\ref{sec:threats} discusses the practical applications and limitations of \watson and the threats to our study.}
Finally, Section~\ref{sec:conclusion} concludes the paper.

\section{Observability for \agentware}
\label{sec:fm_observability}
In this section, we discuss key forms of observability relevant to \agentware, including operational and cognitive observability, and introduce the challenge of recovering reasoning traces in fast-thinking agents.

\subsection{Operational Observability}
\label{sec:fm_observability:operational}
Traditional observability in software systems, rooted in metrics, logs, and traces, has long served to ensure system reliability, diagnose failures, and track performance regressions~\citep{li2022observability}. This paradigm has extended to \agentware, where operational observability focuses on surface-level telemetry such as token usage, latency, and system call outcomes~\citep{data61_agentops}. In agent systems that coordinate multiple model inferences and interact with external components, such observability remains essential for debugging low-level failures and ensuring workflow stability. Several industry and research efforts have contributed observability tooling specifically tailored to \agentware pipelines~\citep{langsmith_url, traceloop_url, wandb_weave_url, langfuse_url, dynatrace_url, lunary_url, phoenix_url, qwak_url, helicone_url}, offering visibility into system bottlenecks and usage patterns.

However, while operational observability can highlight where and when a failure occurs, it provides little insight into why an \llm-powered agent produced a particular response, especially when that response reflects reasoning errors, subtle misunderstandings, or emergent behaviors not grounded in explicit logic. In such cases, simple input-output traces are insufficient for understanding the underlying cognitive processes of \llm-based agents. This gap motivates the need for cognitive observability, a complementary layer of observability that seeks to expose the internal reasoning and decision-making dynamics of agents rather than just their surface behaviors.

\subsection{Cognitive observability}
\label{sec:fm_observability:cognitive}
\del{
Where operational observability is concerned with system performance and resource usage, cognitive observability shifts the focus to higher-level cognitive and linguistic aspects that often cannot be quantified or monitored effectively by traditional tools. Cognitive observability includes three key sub-categories: \textbf{semantic feedback}, \textbf{output integrity}, and \textbf{reasoning trace}. 
The first two---semantic feedback and output integrity---have been explored in recent works~\citep{data61_agentops} and are supported by a growing ecosystem of tools~\citep{langsmith_url, humanloop_url, phoenix_url, whylabs_url, lunary_url}.
Semantic feedback helps gauge the efficacy of outputs and optimize user experience by capturing explicit, implicit, and freeform feedback (e.g., user reactions, ratings, and usage patterns)\citep{nebuly_explicit_implicit_feedback_url, copilot_url}.
Output integrity uses the agent's outputs to gain qualitative insights into their behavior, such as hallucination detection or sentiment analysis\cite{huang2023surveyhallucinationlargelanguage}.
}

\rev{
Where operational observability focuses on system performance and resource usage, we propose the notion of cognitive observability to capture higher-level aspects of an agent's behavior and decision-making that are often not reflected in traditional telemetry. Under this umbrella, cognitive observability can draw on a variety of signals that shed light on how an agent processes information, generates outputs, and its performance is evaluated. Some of these ideas have already been explored in prior work~\citep{data61_agentops}. For example, semantic feedback, which captures explicit, implicit, and freeform user reactions (e.g., ratings, textual feedback, or usage patterns)~\citep{nebuly_explicit_implicit_feedback_url, copilot_url}, and output integrity, which analyzes the agent's outputs to assess qualitative properties such as correctness, hallucinations, or sentiment~\cite{huang2023surveyhallucinationlargelanguage}. Both of these signals, as well as others, have begun to be supported by a growing ecosystem of tools~\citep{langsmith_url, humanloop_url, phoenix_url, whylabs_url, lunary_url}. In contrast, our work emphasizes a less developed aspect of cognitive observability: the recovery of an agent's reasoning trace.
}

\del{
Our work focuses on the third and most underdeveloped sub-category: reasoning trace. This involves exposing the implicit cognitive steps an agent took to arrive at a decision. Understanding an agent's reasoning trace that led to an erroneous decision is crucial for debugging and patching, especially in complex \agentware, where multiple agent's interact and coordinate with both each other and integrated legacy systems. In contrast to traditional software systems, \agentware errors often arise from more abstract or emergent behaviors with no explicit code driving such behavior, meaning that errors can't always be pinpointed through conventional observability tools or debugging techniques.
}

\begin{figure}[!t]
    \centering
    \includegraphics[width=0.91\linewidth]{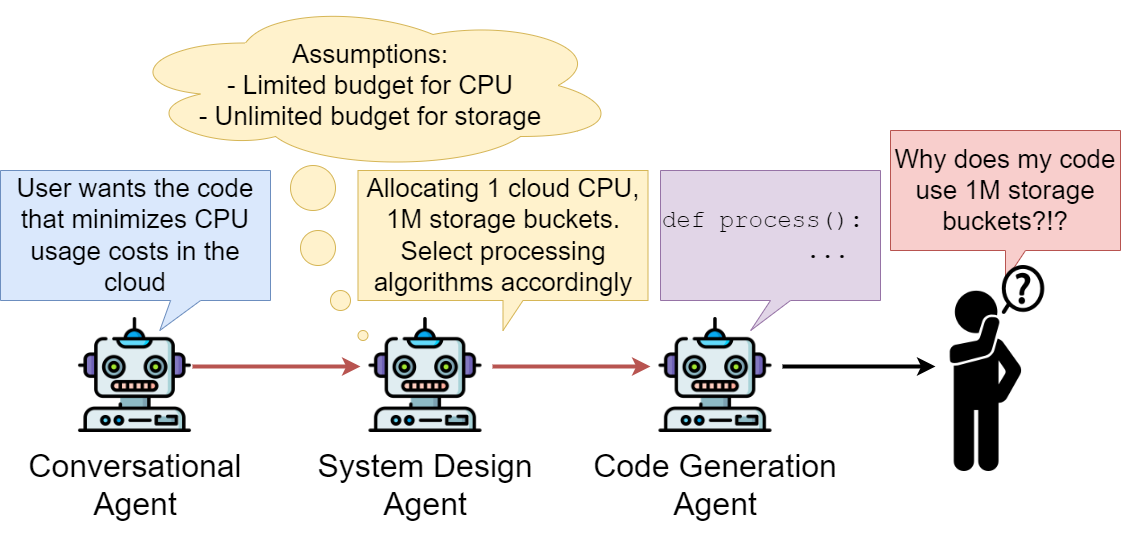}
    \caption{In multi-agent systems, errors made by early agents may only manifest later in the workflow.}
    \label{fig:multi_agent}
\end{figure}

\del{
Additionally, in multi-agent systems, errors made by early agents may only manifest later in the workflow, further complicating the ability to trace back the source of a problem, as demonstrated in Figure~\ref{fig:multi_agent}, where erroneous assumptions of an upstream agent are only manifested later in the workflow. Exposing an agent's reasoning trace is therefore critical for effective debugging, as it enables developers to identify how specific decisions were made, trace the propagation of errors across agents, and systematically resolve failures in complex \agentware systems. However, recovering reasoning traces in fast-thinking \llm agents without affecting the agent's initial output presents a significant challenge, as we discuss in the next section.
}

\rev{
Among the different signals encompassed by cognitive observability, reasoning trace remains the most underdeveloped, which we define as the implicit cognitive steps an agent takes to arrive at a decision. Unlike semantic feedback or output integrity, which can often be observed directly through interactions with users or inspection of outputs, reasoning traces are rarely exposed in fast-thinking \llm agents. Yet, understanding how an agent reached a particular conclusion is crucial for diagnosing and addressing errors, especially in complex \agentware systems where multiple agents interact and coordinate with each other and with external components. In such settings, mistakes made early in the workflow can propagate and only manifest much later, making it difficult to localize their source. As illustrated in Figure~\ref{fig:multi_agent}, erroneous assumptions introduced by one agent may ripple downstream and surface as failures in later stages. Cognitive observability through reasoning traces is therefore essential: it provides developers with a window into the decision-making process, enabling them to identify why an agent behaved as it did, trace the propagation of errors, and more systematically debug failures in complex multi-agent systems.
}

\subsection{Reasoning Trace Recovery in Fast-Thinking Agents}
\label{sec:fm_observability:recoveringreasoning}
\rev{Building on this motivation, we next consider the practical challenge of recovering reasoning traces in fast-thinking \llm agents.} Unlike \lrms~\citep{openai_o1_url, ds_r1}, which explicitly generate chains of thought but incur high latency and cost~\citep{wei2024cot, yao2023treethoughtsdeliberateproblem}, agents driven by fast-thinking \llms do not provide the same level of transparency in their decision-making process. These models generate outputs through a token-by-token sampling process, without producing an internal \quotes{transcript} of how they arrived at a decision. Attempts to recover a reasoning trace (e.g., prompting the model to explain its answer or to \quotes{think step-by-step} using chain-of-thought (CoT) prompting) inevitably change the input and alter the generation process. This interference poses a fundamental challenge to observability: the act of asking for a justification can modify the very behavior we aim to observe. Recent work has shown that CoT prompting can reduce performance on tasks that benefit from fast, intuitive responses. For example,~\citet{liu2024mindstepbystep} found that prompting models to explain their reasoning can significantly reduce accuracy on tasks requiring human intuition or commonsense reasoning, with performance dropping by as much as 36\% compared to direct answering. Similarly,~\citet{su2025dualformercontrollablefastslow} trained a model to switch between \quotes{fast} and \quotes{slow} modes and observed that CoT reasoning led to substantially different completions.

Beyond accuracy differences, there is growing evidence that post-hoc explanations are often unfaithful. \del{The~\citet{anthropic2025reasoningmodels}}\rev{~\citet{chen2025reasoningmodelsdontsay}} demonstrate\del{d} that models exposed to subtle prompt cues would change their answers accordingly, but omit any mention of the cue when asked to justify their output. ~\citet{liao2023aitransparencyagellms} further showed that models often ignore influential prompt features in their explanations, suggesting that self-generated reasoning traces may not reflect the actual decision path taken. Together, these findings highlight a central technical challenge in agent observability: recovering an agent's reasoning trace without modifying its original behavior or inducing hallucinated justifications.

To address these challenges, we introduce a novel approach for retrospectively observing the implicit reasoning of fast-thinking agents without interfering with their original behavior. Rather than prompting the agent to articulate its reasoning during generation, which can distort the output, we instead recover plausible reasoning traces after the fact by mirroring the agent's configuration and constraints. Our focus is not on improving the quality of the output itself, but on faithfully reconstructing the cognitive steps that could have led to it. To ensure these recovered traces are not arbitrary or hallucinated, we incorporate verification mechanisms such as self-consistency checks across multiple reasoning paths and prompt attribution techniques to confirm that the explanations are aligned with the agent's original prompt and completion. This approach forms the foundation of our proposed framework, introduced in the next section.

\section{Watson: A Framework to Observe Reasoning Traces of \llm-powered Agents}
\label{sec:watson}
\begin{figure}[!t]
    \centering
    \includegraphics[width=0.97\linewidth]{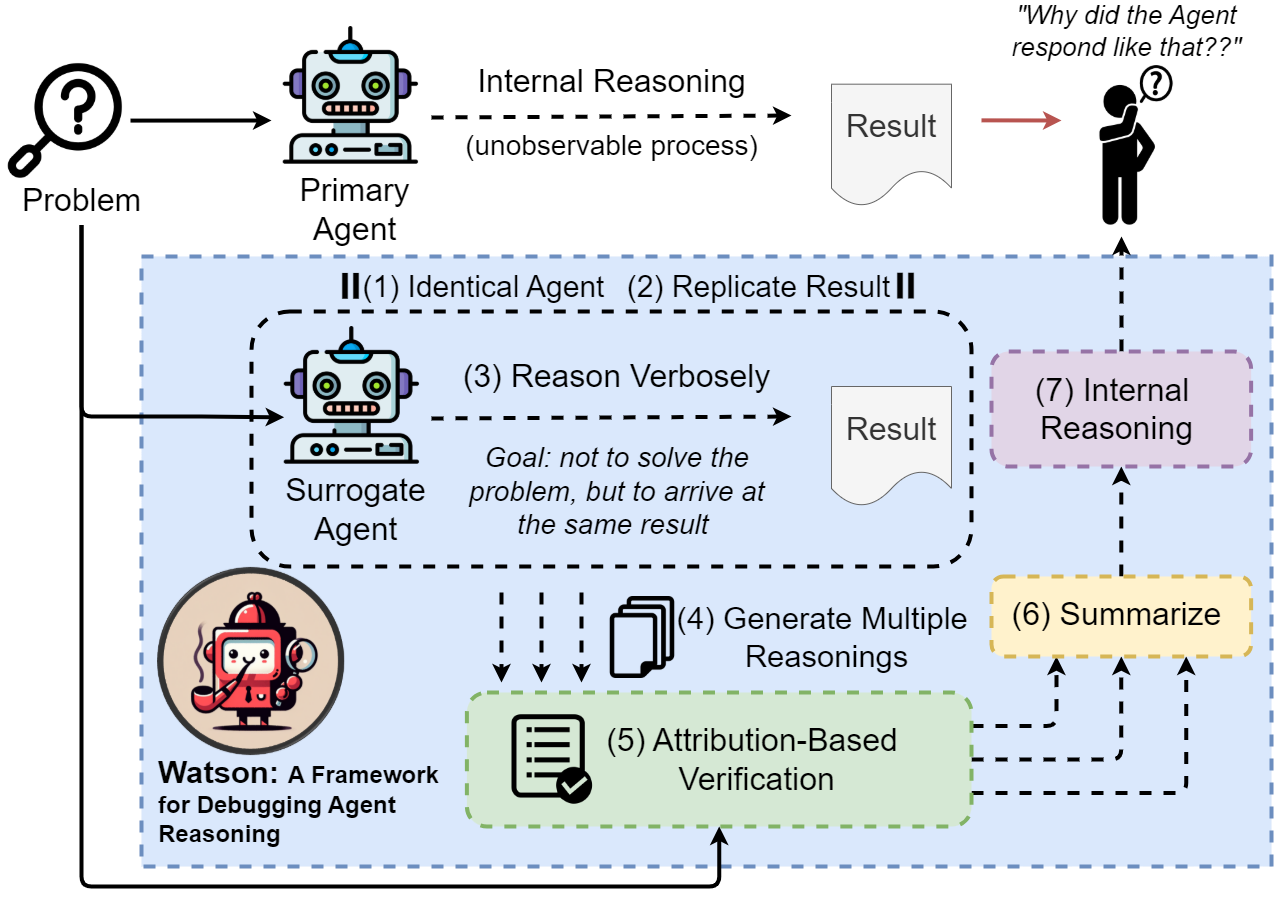}
    \caption{Overview of \watson, a framework to observe the reasoning process of agents without affecting their behavior.}
    \label{fig:watson_overview}
\end{figure}

In this section, we introduce \watson, a framework designed to observe the reasoning traces of \llm-powered agents without interfering with their behavior (Figure~\ref{fig:watson_overview}).

\subsection{Overview}
\label{sec:watson:overview}
In the \watson framework, a \quotes{surrogate agent}~\citep{li2024xai} operates in parallel with a \quotes{primary agent} (i.e., the agent under observation).
The primary agent generates outputs that conform to the constraints and expectations of the system it operates in, and our goal is to observe its behavior without altering it.
In contrast, the surrogate agent reproduces the primary agent’s output while also generating a detailed, step-by-step account of the reasoning process that could have led to that outcome.
This reasoning trace is then verified to assess its fidelity to the primary agent’s actual generation process.
By decoupling reasoning from action, \watson enables developers to recover the implicit reasoning path of the primary agent without affecting its behavior.
The resulting explanations provide valuable insights into the agent’s decision-making process, offering interpretable signals that can be used to debug, evaluate, or improve the agent without altering its operational dynamics.

\watson addresses the challenge of recovering reasoning traces from fast-thinking \llm-powered agents without altering their behavior or introducing unfaithful explanations. It is built around three key ideas. First, the surrogate agent closely mirrors the primary agent to replicate its behavior and ground its reasoning in the same decision process. Second, the surrogate generates reasoning paths linking the input prompt to the \rev{primary agent's} output, providing coherent explanations without affecting the original generation. Third, \watson checks that the surrogate’s reasoning aligns with the actual prompt components that influenced the \rev{primary agent's} output. These validated traces are then summarized into a higher-level \quotes{meta-reasoning}, akin to self-consistency~\citep{wang2023selfconsistencyimproveschainthought}, to capture the agent’s implicit logic. We detail each of these considerations below.

\subsection{Mirroring Configuration of Primary Agent}
\label{sec:watson:mirrorconfiguration}
To accurately recover the primary agent’s implicit reasoning, the surrogate agent must mirror its configuration as closely as possible, most critically in terms of the underlying LLM architecture and decoding parameters. Using the same language model ensures that both agents interpret inputs with equivalent linguistic and reasoning capabilities. Equally important are decoding parameters such as \textit{temperature} and \textit{top\_p}, which control the stochasticity and diversity of generated outputs; even small differences in these settings can produce significant divergence in behavior. While configuration mirroring is necessary for behavioral alignment, it is not sufficient to guarantee output equivalence due to inherent non-determinism in LLM decoding and sensitivity to initialization~\citep{chen2022reproducibledlmodels}. We address this challenge in the final component of the \watson framework (Section~\ref{sec:watson:verifyingreasoning}).

\subsection{Generating Reasoning Paths}
\label{sec:watson:generatingreasoning}
The goal of \watson is to use a surrogate agent to generate the implicit reasoning path the primary agent’s input prompt to its final output. To achieve this, the surrogate agent is given access to the primary agent’s input prompt and the corresponding generated output, and is tasked with producing a reasoning path that potentially bridges the two. However, this introduces a novel design challenge, as most autoregressive \llms are trained for left-to-right generation, predicting the next token based solely on previously seen context. As a result, they lack an inherent ability to generate coherent reasoning that fills in the middle between a fixed prompt and completion.

\begin{figure}[!t]
    \centering
    \includegraphics[width=1.0\linewidth]{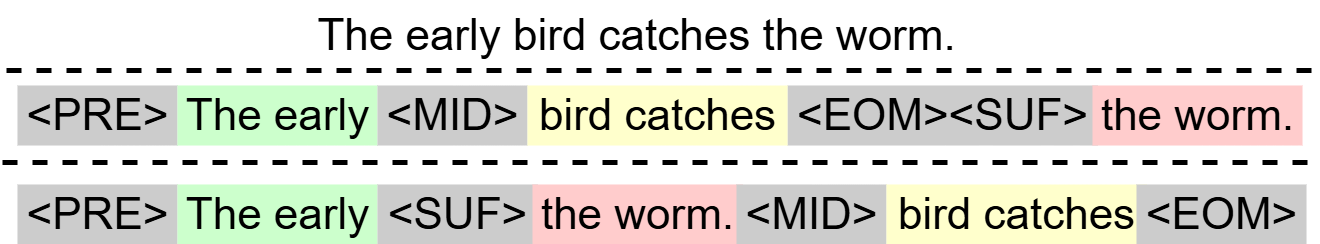}
    \caption{Fill-in-the-Middle (\fim) prompt example. Top: original prompt; middle: with \fim tokens; bottom: prefix-suffix-middle format used for training.}
    \label{fig:fim}
\end{figure}

To address this challenge, we adopt a technique known as \textit{fill-in-the-middle} (\fim)~\citep{bavarian_fim_2022}, which enables decoder-only language models to generate text that fits between a given beginning and end. Originally developed for code completion tasks\citep{izadi_2024_languagemodelscodecompletion, fried2023incodergenerativemodelcode}, \fim allows the model to condition on both a \textit{prefix} and a \textit{suffix} when generating content. The input is segmented using special tokens: \code{<PRE>} for the prefix, \code{<MID>} and \code{<EOM>} to mark the region to be generated, and \code{<SUF>} for the suffix. During training, the model learns to complete the middle section based on the surrounding context, effectively learning to bridge the gap between an input and its corresponding output (Figure~\ref{fig:fim}).

To implement \fim in \watson, the surrogate agent is prompted with the primary agent’s input as the \textit{prefix} and the primary agent’s output as the \textit{suffix}, and is tasked with infilling a plausible reasoning path between them. To better elicit structured reasoning, we lightly augment the original prompt by appending the phrase \quotes{\textit{Let's think step-by-step before arriving at the answer.}} to the prefix, and prepending \quotes{\textit{Therefore, I think the answer is}} to the suffix. This approach leverages the model’s ability to condition on both the beginning and end of a sequence, encouraging it to generate a coherent rationale that links the given input to the observed output. By structuring the prompt in this way, \watson enables the surrogate agent to produce a potential explanation of how the primary agent may have arrived at its response.

While \fim is well suited for generating reasoning paths, its use is limited to \llms that natively support the \fim mechanism and can perform infilling between a given prefix and suffix. In cases where the primary agent relies on an \llm that has not been trained with \fim capabilities, we propose an alternative technique called \quotes{Repetitive Chain-of-Thought} (\repcot). Unlike \fim, \repcot is compatible with any \llm by prompting the surrogate agent with the same input as the primary agent, augmented with an explicit instruction to generate CoT reasoning~\citep{wei2024cot}. To preserve alignment with the primary agent’s behavior, any generations that do not conclude with the same output as the primary agent are discarded. To determine whether the surrogate’s output matches the primary agent’s, we use an \llm-based judge to assess whether the two outputs are semantically equivalent. This step is crucial in \repcot as it ensures that the reasoning path remains consistent with the original decision, capturing only relevant explanations. However, this process may result in higher inference costs due to the need for multiple generations if we must discard generated reasonings that lead to a different outcome from the primary agent's original output.

\subsection{Verification of the Reasoning Consistency}
\label{sec:watson:verifyingreasoning}
To ensure that the surrogate agent’s reasoning accurately reflects the primary agent’s implicit decision process, it is important that the generated reasoning aligns with the original inference of the primary agent that produced the final output. To support this, we first task the surrogate agent with generating a diverse set of potential reasoning paths. This population-based approach helps capture multiple plausible lines of reasoning that could have led to the same answer, offering a broader and more complete view of the primary agent’s underlying thought process.

To address the risk of misaligned explanations, we validate whether each reasoning trace generated by the surrogate agent genuinely reflects the primary agent’s original inference process. This step is crucial, as prior work (see Section~\ref{sec:fm_observability:recoveringreasoning}) has shown that LLMs can omit influential prompt features or hallucinate justifications when asked to explain their outputs~\citep{anthropic2025reasoningmodels, liao2023aitransparencyagellms}. To perform this validation, we apply \promptexp~\citep{promptExp2024}, a framework for prompt explainability that measures the attribution of different prompt components (i.e., how much each part of the input contributed to the final output). The core idea of applying this framework in \watson is to verify that if a particular line of reasoning appears in a surrogate-generated trace, the corresponding component in the original prompt contributed meaningfully to the primary agent’s output.

\rev{
The first step in validating a surrogate-generated reasoning trace is to decompose the input prompt into semantically meaningful components. There are multiple ways to perform this decomposition depending on the type of task being considered. For natural language tasks, the prompt can be split into sentences or clauses; for code tasks, decomposition may be performed by functions or logical blocks; for dialogue tasks, prompt components can correspond to individual speaker turns. Choosing the appropriate granularity is critical because when components are too coarse, subtle but influential elements may be obscured, which can lead to underestimation of their contribution. Conversely, when components are too fine, the resulting attribution scores may be noisy or fragmented, reducing the reliability of the validation process. By default, \watson decomposes prompts into sentences, newlines, and code blocks enclosed in triple backticks. To support more specialized use cases, users may also supply a custom prompt decomposer, enabling task-specific segmentation strategies that go beyond these default heuristics.
}

\del{We begin by decomposing the input prompt into semantically meaningful components (e.g., sentences, code blocks). Then, using \promptexp, e}
\rev{Once the prompt has been decomposed, we use \promptexp to quantify the influence of each component on the primary agent’s output. E}ach component is individually masked to create a series of perturbed inputs, which are then passed to the primary agent to produce a new output. To estimate how much each component influenced the original output, we compare the model’s response to each perturbed input with the original completion. Specifically, we calculate the shift in token probabilities between the two outputs\del{: a larger shift indicates that the masked component played a greater role in shaping the original response.}\rev{, where a larger shift indicates that the masked component played a greater role in shaping the original response.}

\del{These}\rev{The resulting} attribution scores are then aggregated across all components using the method defined in \promptexp\del{, allowing}\rev{. This allows} us to determine which prompt features were most influential in the primary agent’s inference\del{. This}\rev{, and} in turn\del{ enables us}\rev{,} to validate whether a surrogate-generated reasoning trace is grounded in the actual decision factors used by the original model. 

\del{
To perform this validation, we again use an \llm as a judge.
The top- and bottom-ranked prompt components (based on attribution scores) are provided to the judge alongside the original prompt and the surrogate's reasoning. 
The judge is instructed to confirm that the reasoning aligns with the high-attribution components and does not disproportionately emphasize the low-attribution ones. 
If a reasoning trace fails this validation (i.e., it does not reflect the influential prompt features) it is discarded, and new traces are generated until a configurable population size of aligned reasonings is obtained. A simplified concrete example illustrating this reasoning alignment process, including the prompt components, their attribution values, the surrogate reasoning, and validation results, is shown in Table~\ref{tab:reasoning-alignment-example}.
}

\rev{
To perform this validation, we employ an \llm-as-a-judge that explicitly evaluates whether a reasoning trace aligns with the primary agent's influential prompt components. The judge is provided with the original prompt, the surrogate reasoning, and the top-$n$ and bottom-$n$ ranked prompt components based on attribution scores, where $n$ is user-configurable, with a default of 10. We select 10 as it typically covers all components in small- to medium-sized prompts, while for larger prompts it provides a representative subset without overwhelming the judging model with excessive context. In our implementation, the judge is instructed to determine whether the reasoning emphasizes the high-attribution components while not disproportionately focusing on the low-attribution ones, and to respond with a binary output: ``YES'' if the reasoning aligns appropriately or ``NO'' if it does not. Reasoning traces that receive a ``NO'' from the judge are discarded, and new traces are generated until a configurable population size of aligned reasonings is obtained. This setup also allows for flexibility: users can adjust the number of top and bottom components considered or modify the judge instructions to reflect task-specific evaluation criteria.
A simplified example illustrating this reasoning alignment process, including the prompt components, their attribution values, the surrogate reasoning, and the validation results, is shown in Table~\ref{tab:reasoning-alignment-example}.
}

\begin{table}[ht]
\centering
\caption{Illustrative example of reasoning alignment with explicit instructions and nuanced attribution across a multi-part prompt.}
\label{tab:reasoning-alignment-example}
\begin{tabularx}{\columnwidth}{p{1.4cm} X}
\toprule
\textbf{Prompt\newline(Attribution)} & 
\textit{Please evaluate each action below and indicate which is morally wrong. \textbf{(0.10)} \newline
1. I helped a neighbor fix their car. \textbf{(0.10)} \newline
2. I took credit for a colleague's project at work. \textbf{(0.55)} \newline
3. I told a white lie to avoid hurting someone. \textbf{(0.25)}
} \\
\midrule
\textbf{Response} & \textit{The second action is morally wrong} \\
\midrule
\textbf{Attribution\newline Analysis} & The model's output is predominantly influenced by the phrase ``\textit{2. I took credit for a colleague's project at work.}'' with the highest attribution (0.55), and the third phrase being somewhat influential (0.25). The instruction line and first phrase have smaller attributions (0.10 and 0.10). \\
\midrule
\textbf{Reasoning} & \textit{Helping a neighbor fix their car is a kind and altruistic act and therefore not morally wrong. Taking credit for a colleague's project is dishonest and violates ethical standards, making it morally wrong. Telling a white lie to avoid hurting someone's feelings is generally viewed as socially acceptable and thus not morally wrong.} \\
\midrule
\textbf{Validation} & The surrogate reasoning appropriately emphasizes the high-attribution dishonest action and reasonably addresses the other parts, demonstrating good alignment with the model's attribution and output. \\
\bottomrule
\end{tabularx}
\end{table}

Finally, we summarize the reasoning traces that passed verification to extract their mutual threads and recurring ideas, similar to ~\citet{wang2023selfconsistencyimproveschainthought}, who explored identifying shared reasoning across multiple explanations. 
This \del{helps distill}\rev{step produces} a concise explanation that reflects the primary agent’s most likely decision path, while still preserving meaningful variation across the aligned traces. 
By emphasizing points of agreement across the population, the summary highlights reasoning patterns that are both plausible and consistent with the agent’s original behavior. 

To generate this summary, we \del{use}\rev{employ} another \llm acting as a reasoning summarization agent. \del{It}\rev{The agent} is \del{given}\rev{provided with} the full set of verified reasoning traces \rev{as a structured list} and instructed to synthesize them into a single, coherent \del{explanation that captures the most commonly occurring reasoning patterns}\rev{reasoning path}. 
\del{The \llm is prompted to reflect consensus without introducing new ideas, while producing a natural, logically coherent explanation that reads as if it were independently derived rather than a summary.
}
\rev{
The prompt explicitly directs the agent to capture the most frequently occurring reasoning patterns, maintain logical coherence, avoid contradictions, and produce a natural, standalone explanation without introducing any new ideas or meta-summary language. This approach ensures that the resulting summary reflects consensus among the verified traces while remaining faithful to the original reasoning processes.
}

\section{Case Study}
\label{sec:evaluation}
This section presents a case study illustrating how the \watson framework enables semantic observability in \llm-powered agentic systems. Rather than evaluating or extending \watson itself, we use it as a diagnostic tool—analogous to a debugger in traditional software development. We begin with an example of \watson-generated reasoning and a preliminary manual study in which a developer uses reasoning traces to diagnose real-world failures in the \acr system. This demonstrates the interpretability and practical utility of \watson’s outputs for fault localization.

We then show how \watson-generated can support automated debugging pipelines across two domains: multiple-choice question answering and autonomous software engineering. Together, these studies demonstrate that semantic observability enables both human and agentic insight into model reasoning. The remainder of this section outlines the purpose of the case study, describes the experimental setup, and presents results highlighting the value of incorporating reasoning traces into agent workflows.

\subsection{Purpose and Positioning}
\label{sec:evaluation:purpose}
This case study demonstrates the practical utility of the \watson framework by applying it to agentic systems and using the \watson-generated reasoning traces to both manually and automatically debug the performance of a \llm-powered agent. Importantly, this case study does not extend the \watson framework itself. Rather, it builds on top of \watson as a semantic observability tool, illustrating how the reasoning traces it produces can be used both by human developers and in an external, automated debugging pipeline.

The goal of this case study is not to directly enhance the reasoning capabilities of the underlying \llm or to optimize the accuracy of the primary agent. Instead, we draw a parallel to traditional software debugging workflows, where tools such as the GNU Debugger (GDB)~\citep{gdb} are employed to observe and diagnose faults in program execution, without being expected to fix or improve the software themselves. Similarly, \watson enables the observation of an \llms latent reasoning traces, which can then be analyzed or acted upon by a human or another agent.

To support this demonstration, we first conducted a preliminary manual debugging study, in which we analyze \watson-generated reasoning traces to identify failure modes in agent behavior. 
In our analysis, the traces revealed latent reasoning errors and misaligned assumptions that were not evident from the final answers alone. Building on these findings, we introduce an automated debugging agent that consumes the same reasoning traces to detect and intervene on faulty reasoning paths. 
This agent operates independently of both the primary and surrogate agents, demonstrating how \watson facilitates semantic observability without modifying the underlying model or agents.

Although we report metrics such as pass@$k$ improvement to quantify the outcomes of this automated debugging process, these results should be interpreted primarily as evidence of the practical utility of semantic observability. Rather than evaluating \watson{}’s ability to directly optimize model reasoning, our goal is to illustrate how external systems, either human or automated, can effectively leverage reasoning traces produced by \watson to enhance transparency and support debugging workflows.

\subsection{Task Selection and Study Design}
\label{sec:evaluation:design}
We evaluate the utility of \watson on both non-agentic and agentic tasks to demonstrate its generalizability and effectiveness across domains. The first task involves the \mmlu benchmark, a diverse set of multiple-choice questions spanning subjects such as professional law, high school psychology, and moral reasoning. From the full set of 14,042 questions, we randomly sample three subsets totaling 374 questions each, which corresponds to the minimum required for statistical significance at a 95\% confidence level with a 5\% margin of error. These tasks serve as a controlled environment for observing \llm behavior outside of agentic workflows.

As a first step, we present a qualitative example from \mmlu to illustrate the kinds of reasoning \watson can produce. In Listing~\ref{lst:watson-law-reasoning}, we show an incorrectly answered question from the professional law category, where the agent selects option \textit{B} by focusing on the offer’s duration, rather than the correct option \textit{A}, as well as the corresponding reasoning trace generated by \watson. The trace (1) accurately reflects the primary agent’s chosen answer, (2) considers the merits of alternative options, and (3) articulates a coherent rationale for the agent’s decision. This example highlights \watson’s ability to produce detailed and logically grounded reasoning even in static, non-agentic tasks.

\begin{lstlisting}[
    language=,
    basicstyle=\fontbetween\ttfamily,
    caption={Example MMLU benchmark question, the initial incorrect answer, and the associated reasoning trace generated by \watson},
    label={lst:watson-law-reasoning},
]
You are a general knowledge quiz master. You are going to answer a question from a quiz. Only respond with the correct answer.
[Question] A merchant's irrevocable written offer (firm offer) to sell goods 
[Options] 
A) Must be separately signed if the offeree supplies a form contract containing the offer.
B) Is valid for three months.
C) Is nonassignable.
D) Can not exceed a three-month duration even if consideration is given.
[Answer] B) Is valid for three months.
[Reasoning] The question is asking for the characteristics of a merchant's irrevocable written offer to sell goods.
Option A mentions that the offer must be separately signed if the offeree supplies a form contract containing the offer. This statement does not directly answer what is being asked in the question.
Option B states that the offer is valid for three months. This seems like it could be a possible answer, as it specifies a duration for which an offer can be considered valid.
Option C states that the firm offer is nonassignable. While this may or may not be true, it does not pertain to the characteristics of a merchant's irrevocable written offer.
Lastly, option D states that such an offer cannot exceed three months even with consideration given. Again, while this may or may not be true, it does not directly address what makes up a merchant's irrevocable written offer to sell goods.
Based on these options and considering all factors involved in answering this question correctly, I believe option B - Is valid for three months - would make sense as an appropriate response because: (1) It specifically addresses one characteristic of such an offer (validity period), (2) It doesn't mention anything outside of what was asked, (3) The other options don't seem relevant enough
\end{lstlisting}

\begin{figure*}[!t]
\begin{lstlisting}[
    language=,
    basicstyle=\fontbetween\ttfamily,
    caption={The \watson-generated reasoning for instance \code{django\_\_django-13401}, highlighting key reasoning steps. The agent incorrectly focuses on \code{model\_inheritance/tests.py}, which contains subclasses \code{B} and \code{C}, and proposes changes there, rather than targeting the actual source of the bug.},
    label={lst:django_13401_summary},
    aboveskip=0pt,
    belowskip=0pt,
]
The common points across the provided explanations can be summarized as follows:
1. **Identified Bug Location**: All reasonings consistently highlight that the issue is centered around the (*@\colorbox{lightblue}{\texttt{model\_inheritance/tests.py} file}@*), which contains classes A, B, and C. This file is repeatedly mentioned as critical to understanding where potential bugs may lie.
2. **Focus on Classes B and C**: (*@\colorbox{lightblue}{There is a shared emphasis on examining classes B and C}@*) due to their inheritance from class A, suggesting that (*@\colorbox{lightblue}{any issues likely originate from these subclasses}@*) or the abstract base class.
3. **Method Investigations**: Several explanations point out specific methods (`__eq__`, `__lt__`, and `__hash__`) in relation to fields within these classes, indicating a need to analyze how equality comparisons are implemented for instances of these models.
4. **Sufficient Context Established**: Multiple reasonings express confidence that enough context has been gathered regarding relevant files and classes without needing further information at this stage.
5. **Search Strategies Proposed**: Some responses suggest utilizing search functions like `search_method_in_class()` or looking for similar cases using other search tools to uncover additional insights about how fields are compared across different models.
\end{lstlisting}
\end{figure*}

\begin{figure*}[!t]
\begin{lstlisting}[
    language=,
    basicstyle=\fontbetween\ttfamily,
    caption={\rev{Segment of \acr output for instance \code{django\_\_django-13401} with a reasoning hint. The hint guides \acr to the correct patch in \code{django/db/models/fields/\_\_init\_\_.py} and to rule out \code{tests.py} as the source.}},
    label={lst:django_13401_new_output},
    columns=flexible,
    aboveskip=0pt,
    belowskip=0pt,
]
[ASSISTANT] The issue reported is related to the equality comparison of two fields in different models. The author found that if they pull the fields of two models into a shared set, one will be de-duplicated away, because they compare as equal. The root of the issue is that they compare equal, as fields only consider self.creation_counter when comparing for equality. So, we need to adjust __eq__ so that if the field.model is different, they will compare unequal. Similarly, it is probably wise to adjust __hash__ and __lt__ to match. When adjusting __lt__, it may be wise to order first by self.creation_counter so that cases not affected by this equality collision won't be re-ordered. In the experimental branch, there was one test that broke if ordered them by model first. 
(*@\colorbox{lightblue}{The fix for this issue is not located in any \texttt{tests.py} files}@*), although they are a good place to start to get more context on the issue. We also found that the code snippets we searched for are mostly located in the file (*@\colorbox{lightblue}{\texttt{django/db/models/fields/\_\_init\_\_.py}}@*), which makes sense as this is where the Field class is defined. We can also see that the class A is inherited by classes B and C, and B and C are inherited by other classes in the codebase. This could be the reason why the issue is not located in any `tests.py` files, as it may be related to the inheritance of classes. Based on the context we have gathered, we can now start writing patches to solve the issue.
\end{lstlisting}
\end{figure*}

To evaluate \watson in more complex, real-world settings, we incorporate \acr, an autonomous software engineering agent that resolves GitHub-style issues by analyzing and modifying source code. We apply \acr to \swebenchlite, a benchmark of 300 issues derived from the broader \swebench dataset~\citep{jimenez2024swebenchlanguagemodelsresolve}. \acr operates in two main phases: first, it retrieves and synthesizes relevant context from the codebase to identify likely faulty regions; second, it generates code patches to fix the issue. Because overall performance hinges on the agent’s initial code localization step, our evaluation focuses exclusively on this first, opaque phase.

\rev{To assess generalizability across agent scaffolds, we additionally evaluate the \oh on the same \swebenchlite benchmark. This allows us to observe whether \watson can similarly capture latent reasoning traces and guide debugging across different agent architectures and underlying models.}

To gain qualitative insight into \watson’s effectiveness in this setting, we conduct a manual debugging analysis on two failure cases from the \acr evaluation. A human developer reviews the issue report, the original \acr output, and the \watson-generated reasoning trace (Section~\ref{sec:watson}) to diagnose potential missteps in the agent’s reasoning. Based on these insights, the developer then crafts a targeted decision-refining hint, which is fed, along with the original prompt, back into \acr to reattempt the issue. This manual study helps assess whether \watson's traces can meaningfully aid human understanding and correction of complex agent behavior.

Building on these qualitative findings, we conduct a full-scale automated study using dedicated debugging agents on \del{both \mmlu and \acr}\rev{\mmlu, \acr, and \oh}. In \del{both cases}\rev{each case}, we generate hints from the reasoning traces and inject them into the agent’s prompt to guide its decision-making. For \mmlu, we evaluate the impact of these hints on answer accuracy, while for \rev{both} \acr \rev{and \oh}, we assess their effect on localization performance across the entire benchmark.

\subsection{Manual Debugging with Reasoning Traces}
\label{sec:evaluation:manual}
To explore the practical utility of \watson in real-world debugging scenarios, we conducted a manual analysis using two failure cases from the \acr evaluation set (i.e., cases where the incorrect code files were identified by the first agent as buggy locations). 
For each failure case, we began by presenting the corresponding issue report and the output of running \acr on the task to a human developer, along with the associated reasoning trace generated by \watson (Section~\ref{sec:watson}). The developer then attempted to infer which aspects of the issue were emphasized or overlooked by the agent and whether these misalignments could plausibly explain the failure. Based on this analysis, we constructed a decision-refining hint and provided it, alongside the original input prompt, to \acr in an attempt to re-solve the issue. 

When executing these manual cases, we use the \fim configuration of \watson, and use \gptthreefiveturboinstruct as the underlying model, as it is the only OpenAI model that explicitly supports the \fim capabilities required. In this setup, both the primary and surrogate agents share the same model and decoding parameters to ensure consistency. \rev{For reasoning validation, we employ the default prompt decomposer and set $n=10$, meaning that the judge considers up to the 10 most- and 10 least-attributed prompt components to verify that the surrogate reasoning aligns with the aspects the agent attended to.} We generate a minimum of 10 reasoning traces per task. To summarize the resulting reasoning populations and perform related analysis tasks, we use OpenAI’s \gptfouromini model, selected for its improved reasoning quality and reliability.

\begin{figure*}[!t]
\begin{lstlisting}[
    language=,
    basicstyle=\fontbetween\ttfamily,
    caption={\rev{\watson-generated reasoning for instance \code{django\_\_django-14238}, showing focus on \code{example/core/models.py} while overlooking the issue’s reference to \code{\_\_subclasscheck\_\_}, absent from all explanations.}},
    label={lst:django_14238_summary},
    aboveskip=0pt,
    belowskip=0pt,
]
**Common Points Across Reasonings:**
1. **Focus on "MyModel" and `__init__` Method**: Multiple explanations consistently mention the class "MyModel" and its constructor method `__init__`, indicating that they are central to identifying the bug locations.
2. **DEFAULT_AUTO_FIELD Issues**: There is a recurring theme regarding problems with `DEFAULT_AUTO_FIELD`, suggesting it is linked to the bugs being analyzed.
3. **File Location**: (*@\colorbox{lightblue}{The file "example/core/models.py" is frequently identified as a potential}@*) (*@\colorbox{lightblue}{source of issues, reinforcing its importance in diagnosing the problem.}@*)
4. **Need for Additional Context**: Several reasonings highlight whether more context or information from other files/methods might be necessary, reflecting uncertainty about fully understanding all aspects of the issue.
\end{lstlisting}
\end{figure*}

\subsubsection{Instance django\_\_django-13401}\fnu{https://code.djangoproject.com/ticket/31750} 
This issue, reported in the \django~\citep{django_url} web framework, concerns a bug where attribute fields from different models—each inheriting from the same parent model—are incorrectly treated as equal. The problem arises because field equality is determined solely by the \code{self.creation\_counter} attribute. As a result, fields from models \code{B} and \code{C}, both of which inherit from model \code{A}, are considered equal despite belonging to different subclasses. This leads to unexpected behavior—for example, when fields from models \code{B} and \code{C} are added to a set, one may be erroneously deduplicated. The issue reporter proposes modifying the \code{\_\_eq\_\_}, \code{\_\_hash\_\_}, and \code{\_\_lt\_\_} methods to ensure that fields from different parent models are correctly treated as unequal.

\smallskip \noindent \textbf{Output of Reasoning}: The reasoning generated from \watson for \code{django\_\_django-13401}, shown in Listing~\ref{lst:django_13401_summary}, suggests the agent is focusing on the test files, specifically \code{model\_inheritance/tests.py}, which is not the code location for where we expect the solution patch to be applied. Following this incorrect decision pathway, \acr attempts to modify subclasses \code{B} and \code{C}, which are contained in \code{model\_inheritance/tests.py}, as the fix for the encountered issue.

\smallskip \noindent \textbf{After Fixing with Hints}: Based on the \watson-generated reasoning, we add the following hint at the end of the issue report and rerun \acr.

\begin{myframe}[width=\linewidth, top=0pt,bottom=0pt,left=5pt,right=5pt,arc=8pt,auto outer arc]
\small\textit{The fix for this issue is not located in any ``tests.py'' files, although they are a good place to start to get more context on the issue.}
\end{myframe}

With this hint, \acr arrives at the correct file location to be patched, while specifically acknowledging that the fix is not located in any test files (Listing~\ref{lst:django_13401_new_output}).

\subsubsection{Instance \code{django\_\_django-14238}}\fnu{https://code.djangoproject.com/ticket/32620}
This issue is reported in the \django framework and describes an application crash during model initialization when \code{DEFAULT\_AUTO\_FIELD} is set to a custom subclass of \code{BigAutoField}.
The issue report includes a traceback demonstrating a \code{ValueError} requiring that the custom primary key class must subclass \code{AutoField}. 
The issue stems from a failure in \code{AutoFieldMeta.\_\_subclasscheck\_\_}, which does not recognize valid subclasses of \code{BigAutoField} or \code{SmallAutoField}. 
The issue report includes a suggested fix to modify the subclass check to include these subclasses in the \code{\_subclasses} property.

\smallskip \noindent \textbf{Output of Reasoning}: The reasoning generated from \watson for \code{django\_\_django-14238}, shown in Listing~\ref{lst:django_14238_summary}, suggests that the agent is placing its focus on the files in the \code{example} directory. However, upon closer inspection of the original issue report, it is explicitly mentioned that the fix can be applied to the \code{\_\_subclasscheck\_\_} method, which does not show up at all in the reasonings, indicating that the agent is ignoring this information.

\smallskip \noindent \textbf{After Fixing with Hints}: Based on the reasonings, we decide to add the following hint at the end of the issue report and rerun \acr.

\begin{myframe}[width=\linewidth, top=0pt,bottom=0pt,left=5pt,right=5pt,arc=10pt,auto outer arc]
\small\textit{The fix for this issue would not be located in the `example` directory. Make sure you focus on the `\_\_subclasscheck\_\_` method.}
\end{myframe}

With this hint, \acr retrieves the correct contextual information related to \code{\_\_subclasscheck\_\_} and arrives at the correct file location to be patched (Listing~\ref{lst:django_14238_new_output}).

\begin{figure*}[!t]
\begin{lstlisting}[
    language=,
    basicstyle=\fontbetween\ttfamily,
    caption={\rev{Segment of \acr output for instance \code{django\_\_django-14238} with a reasoning hint, enabling \acr retrieve context on \code{\_\_subclasscheck\_\_} and locate the correct patch in \code{django/db/models/fields/\_\_init\_\_.py}, avoiding prior focus on the \code{example} directory.}},
    label={lst:django_14238_new_output},
    columns=flexible,
    aboveskip=0pt,
    belowskip=0pt,
]
[ASSISTANT] Based on the search results, we can see that there are 12 classes with the name MyModel in the codebase, but (*@\colorbox{lightblue}{none of them are located in the "example" directory}@*). This means that the issue is not related to the "example" project.
We also found 1 method with the name __subclasscheck__, which is located in the django/db/models/fields/__init__.py file. This method is inherited from the AutoFieldMeta class and is used to check if a subclass is a subclass of AutoField.
Additionally, we found 3 snippets containing `DEFAULT_AUTO_FIELD =` in the codebase. These snippets are located in the tests/test_sqlite.py, tests/check_framework/test_model_checks.py, and django/conf/global_settings.py files. These snippets are used to set the default primary key field type in the codebase.
Finally, we found 2 snippets containing `__subclasscheck__` in the codebase, both of which are located in the django/db/models/fields/__init__.py file. These snippets are used to check if a given subclass is a subclass of AutoField.
Based on this context, (*@\colorbox{lightblue}{we can conclude that the issue is related to the django/db/models/fields/}@*) (*@\colorbox{lightblue}{\_\_init\_\_.py file}@*), specifically the (*@\colorbox{lightblue}{AutoFieldMeta class and its \_\_subclasscheck\_\_ method.}@*) The fix for this issue would not be located in the "example" project, but rather in the django project itself.
\end{lstlisting}
\end{figure*}

\begin{table}[ht]
\centering
\caption{\rev{Pass@1 / Pass@3 Accuracy Across Evaluation Runs on \mmlu\ (374 questions per run). Configurations: RO-SR = Response-only Self-Reflection, WA-SR = Watson-Aided Self-Reflection.}}
\label{tab:mmlu_results}
\begin{tabular}{lccc}
\toprule
\textbf{Configuration} & \textbf{Run 1} & \textbf{Run 2} & \textbf{Run 3} \\
\midrule
RO-SR (FIM)   & 59.4\% / 68.2\% & 59.4\% / 69.8\% & 60.7\% / 71.1\% \\
WA-SR (FIM)    & 58.8\% / 66.8\% & 61.0\% / 71.4\% & 63.3\% / 73.5\% \\
\midrule
RO-SR (RepCoT) & 63.6\% / 70.9\% & 63.1\% / 71.4\% & 66.3\% / 74.3\% \\
WA-SR (RepCoT)  & 61.2\% / 73.0\% & 66.0\% / 74.3\% & 67.9\% / 77.2\% \\
\bottomrule
\end{tabular}
\end{table}

\subsection{Automated Debugging Pipeline}
\label{sec:evaluation:automated}
Building on the insights gained from the manual debugging study, we design a two-stage automated evaluation pipeline to demonstrate how \watson-generated reasoning traces can be effectively leveraged in downstream debugging and decision-making processes. Rather than aiming to directly improve agent performance, this pipeline seeks to showcase the practical utility of semantic observability by comparing system behavior with and without \watson’s reasoning traces.

\rev{To clearly isolate the contribution of each component, we generate guiding hints for the agent under two distinct cases: (1) using \watson-generated reasoning traces along with the agent’s original answer, and (2) using only the agent’s original answer without any reasoning traces. This second case serves as an explicit ablation run, allowing us to measure the added value of incorporating the reasoning traces into the hint.}

We begin by establishing a baseline through an ablation study. Each agentic system is first executed in its standard form (i.e., without the involvement of \watson) to produce an initial answer. For \mmlu, this answer corresponds to the selected option for a benchmark question; for \acr\rev{ and \oh}, it refers to the code location identified as faulty. We then simulate a self-reflection~\cite{2023reflexion} process by prompting a separate \llm to evaluate whether the produced answer is reasonable in the context of the task (i.e., the multiple-choice question or the issue report). The reflection is performed without access to any reasoning traces, relying solely on the answer and task description. The agent is subsequently rerun on the task, with the self-reflection output injected into its prompt as a guiding hint. The correctness of the newly generated answer is then evaluated, and the result constitutes the baseline performance.

Next, we repeat the process with \watson applied to the same set of tasks. In this configuration, the surrogate agent generates a reasoning trace that explains how the primary agent arrived at its answer. This trace is incorporated into the self-reflection stage, enabling the reflection \llm to critique both the final answer and the reasoning path that led to it. As before, the resulting reflection is provided to the agent, which reruns the task with the added guidance. We then assess the correctness of this revised output. \rev{In summary, for each task, we generate two hints: one that leverages \watson’s reasoning trace (the \quotes{hint from reasoning+answer} case) and one that does not (the ablation \quotes{hint from answer only} case).}

\del{For each task, we run two versions of the evaluation pipeline: one without \watson enabled, and one with \watson enabled.} \rev{Building on this setup, we systematically evaluate each task under both ablation (without \watson) and \watson-enabled conditions. For the MMLU evaluation, the \watson-enabled pipeline is tested under both configurations of the \watson framework: \fim, using \gptthreefiveturboinstruct as the underlying language model, and \repcot, using \gptthreefiveturbo-0125, which offers chat-based interaction while maintaining close behavioral similarity to \gptthreefiveturboinstruct. For the \acr experiment, we use the \fim configuration with \gptthreefiveturboinstruct, while for the \oh experiment, we use the \repcot configuration with Qwen3-8B.} In both configurations, the primary and surrogate agents still share the same underlying model and are executed using identical decoding parameters\del{, and w}\rev{. For reasoning validation, we also employ the default prompt decomposer and set $n=10$, such that the judge examines the 10 highest- and lowest-attributed prompt components to verify alignment between the surrogate reasoning and the agent’s attention. W}e generate a minimum population size of 10 reasonings. Finally, tasks within the evaluation pipeline (e.g., summarization of reasoning populations, hint generation, and grading outputs) are performed using OpenAI’s \gptfouromini model, selected for its improved reasoning quality and reliability in downstream analysis tasks. \rev{We compute \texttt{pass@1} and \texttt{pass@3} for \mmlu, and \texttt{pass@1} for \acr and \oh.}

\del{To quantify performance, we compute \texttt{pass@1} and \texttt{pass@3} metrics for both the baseline and \watson-enabled runs. These scores indicate the proportion of tasks for which the correct answer appears in the top one or top three attempts, respectively. Improvements in these metrics are used to demonstrate how semantic observability via \watson can enhance downstream agent behavior in an automated debugging context. }

\subsection{Evaluation Results}
\label{sec:evaluation:results}
This section presents the empirical results of applying the \watson framework across \del{\mmlu and \acr}\rev{\mmlu, \acr, and \oh}. Our evaluation compares agent performance under two conditions: (1) a self-reflection pipeline without access to \watson reasoning traces, and (2) a self-reflection pipeline enhanced with \watson-generated reasoning traces. \del{We report quantitative results for both reasoning generation configurations supported by \watson (i.e., \fim and \repcot).}

\subsubsection{\mmlu Results}
Across three statistically significant evaluation runs on the \mmlu benchmark (Section~\ref{sec:evaluation:design}), we investigate whether incorporating \watson-generated reasoning traces supports the surrogate agent's self-reflection process. Rather than demonstrating uniform improvements in accuracy, our results, shown in Table~\ref{tab:mmlu_results}, highlight nuanced shifts in performance metrics that suggest the potential value of incorporating \watson-generated reasoning into the reflection evaluation process. These effects are observed across both the \fim and \repcot configurations of \watson.

In the \fim setting, the baseline Pass@1 accuracy across the three runs ranges from 59.4\% to 60.7\%, while incorporating reasoning traces increases this to a range of 58.8\% to 63.4\%. Although Pass@1 performance shows marginal variance in the first experiment, the gains become more substantial in subsequent runs. The improvements are more consistently observed in Pass@3 accuracy, which increases from 68.2\%-71.1\% in the baseline to 66.8\%–73.5\% with reasoning, with each of the three runs demonstrating improvement. \rev{The $\chi^2$ test shows no significant differences (Pass@1 $p$=0.88,0.65,0.45; Pass@3 $p$=0.50,0.63,0.46), and overlapping 95\% CIs indicate gains are trends, not definitive improvements.}

\del{The \repcot configuration shows even stronger gains. B}\rev{For the \repcot configuration, the b}aseline Pass@1 scores range from 63.1\% to 66.3\%, while the inclusion of reasoning increases them to 61.2\%–67.9\%, with the largest improvement observed in the third run. More prominently, Pass@3 accuracy improves from 70.9–74.3\% (baseline) to a range of 73.0–77.2\% when \watson-generated reasoning is used. These results indicate that \watson's explicit reasoning generation capabilities enhance the agent's self-reflective judgment, especially in settings where structured reasoning like RepCoT is applied. \rev{As with the \fim setting, the $\chi^2$ test shows no significant differences for \repcot (Pass@1 $p$=0.50,0.40,0.64; Pass@3 $p$=0.52,0.37,0.35), and overlapping 95\% CIs indicate gains are trends, not definitive improvements.} However, the gains in Pass@3 suggest that reasoning traces enable the agent to better recognize partial correctness and near-miss solutions, improving its overall evaluative reliability.

\subsubsection{\acr \rev{and \oh} Results}
\del{
Evaluating \acr presents a unique challenge due to its iterative code localization process, which often approaches the maximum context window limit of the underlying language model. As a result, some runs terminate prematurely without identifying a buggy code location. To account for this, we categorize each evaluation attempt into one of three outcomes: \textit{invalid}, \textit{incorrect}, and \textit{correct}. A case is labeled \textit{invalid} if \acr fails to identify any buggy location before hitting the context limit. If \acr does identify a location but it is incorrect, the case is marked \textit{incorrect}. If the identified location is correct, it is labeled \textit{correct}.
}
\del{
To mitigate the risk of premature failure, we adopt a retry mechanism: for each task, \acr is executed up to three times in an effort to reach either a correct or incorrect state. This strategy aims to reduce the number of discarded tasks due to invalid runs. The same retry logic is applied when evaluating both the baseline (no \watson involvement) and the \watson-enhanced self-reflection pipelines. If all three attempts result in invalid outcomes, the task is excluded from further analysis.
}
\del{
This categorization and retry strategy allow us to more reliably evaluate the impact of \watson-generated reasoning on downstream debugging accuracy in code localization tasks, while controlling for the practical limitations of current \llm{}'s context windows. Given the computational cost and time required to execute multiple retries per sample, we focus our analysis on the pass@1 metric to ensure a tractable yet meaningful evaluation.
}
\rev{
For both \acr and \oh, we focus our evaluation on tasks where the agent successfully identified a buggy location. Each case is categorized as \textit{correct} if the located code region matches the ground truth and \textit{incorrect} otherwise; cases where the agent fails to identify any location are excluded from analysis. This approach allows us to assess the impact of \watson-generated reasoning on downstream debugging accuracy without being confounded by runs that terminate prematurely. Given the computational cost and practical constraints, we report \texttt{pass@1} for both agents to provide a tractable yet meaningful measure evaluation.
}

\del{
The initial run of \acr on \swebenchlite resulted in 103 \textit{valid} tasks, meaning that the agent was able to identify potential defective code locations without exceeding the context window limit. Using these valid samples, we evaluated two self-reflection pipelines. The baseline pipeline achieved a Pass@1 of 71 out of 103, while the \watson-enabled self-reflection pipeline improved slightly, achieving a Pass@1 of 75 out of 103. This indicates a modest gain in localization accuracy when incorporating \watson-generated reasoning (Table~\ref{tab:acr_results}).
}

\rev{
Across the two agentic experiments on \swebenchlite, we focus on tasks where the agent successfully identified potential buggy code locations. For \acr, 103 valid tasks were analyzed: the baseline self-reflection pipeline achieved a Pass@1 of 71, while the \watson-enabled pipeline reached 75. For \oh, 57 valid tasks were considered, with a baseline Pass@1 of 33 and a \watson-enabled Pass@1 of 34 (Table~\ref{tab:acr_results}). In both cases, the slight improvements observed with \watson-generated reasoning were not statistically significant ($p>0.05$), indicating trends rather than definitive gains in localization accuracy.
}

\begin{table}[ht]
\centering
\caption{Pass@1 Accuracy of Self-Reflection Pipelines on Valid \acr and \oh Samples (\swebenchlite). RO-SR = Response-only Self-Reflection, WA-SR = Watson-Aided Self-Reflection.}
\label{tab:acr_results}
\begin{tabular}{lcccc}
\toprule
\textbf{Configuration} & \textbf{Valid Samples} & \textbf{RO-SR} & \textbf{WA-SR} \\
\midrule
\acr   & 103 & 71 (68.9\%) & 75 (72.8\%) \\
\oh &  57 & 33 (57.9\%) & 34 (59.6\%) \\
\bottomrule
\end{tabular}
\end{table}

\begin{revision}
\subsection{Comparison with Native Reasoning Models}
To further evaluate the quality of \watson’s post-hoc reasoning traces, we compared them against traces produced natively by a reasoning-capable large language model. We selected Qwen3-8B, which provides a \quotes{thinking} mode to explicitly generate reasoning alongside answers. Using a statistically representative MMLU sample from Section~\ref{sec:evaluation:design}, we ran the model twice per example: once with \quotes{thinking} enabled to collect native reasoning, and once with reasoning disabled to obtain only the final answer. In the latter case, we applied \watson to the input–output pair to generate post-hoc reasoning traces using the \repcot configuration, as Qwen3-8B does not support \fim.

Each Watson-generated trace was paired with its corresponding native Qwen3-8B trace. We used an \llm-as-a-judge to assess whether the two traces were semantically equivalent, focusing on whether they followed the same logical progression, invoked similar assumptions, and reached conclusions through comparable reasoning, while ignoring superficial wording differences.

For Qwen3-8B, we compared \watson{}’s post-hoc reasoning traces with the model’s natively generated reasoning traces, conditioned on whether both runs produced the same final answer. Across the sampled MMLU tasks, the two runs produced the same answer in 307 cases (76.2\%) and diverged in 67 cases (16.6\%). When the answers matched, the reasoning traces were semantically equivalent in 212 cases (69.1\%) and diverged in 95 cases (30.9\%). Conversely, when the answers differed, the reasoning traces almost always diverged (64 cases, 95.5\%), with only a small number of coincidental overlaps (3 cases, 4.5\%, $p < 10^{-20}$, Fisher’s exact test). While the divergence observed in roughly one-third of the matching-answer cases might initially appear as a limitation, it could in fact be advantageous: recent work has shown that natively generated reasoning traces are not always faithful to the model’s actual inference process, often reflecting superficial rationalizations rather than genuine decision pathways~\citep{chen2025reasoningmodelsdontsay, barez2025chainofthoughtisnotexplainability}. By contrast, \watson enforces consistency with the model’s saliency during inference, allowing it to surface potential discrepancies between an agent’s stated reasoning and its underlying behaviour. In this sense, divergences revealed by \watson may highlight cases where post-hoc reasoning provides a truer window into the model’s decision-making, offering a valuable diagnostic signal for debugging and analysis.
\end{revision}

\subsection{Summary}
This case study demonstrates the utility of \watson as a semantic observability framework for supporting both manual and automated debugging of \llm-powered agents. Across static (\mmlu) and dynamic (\acr \rev{and \oh}) tasks, we find that \watson-generated reasoning traces consistently surface actionable signals that aid fault localization and enable targeted interventions. Our findings reinforce that \watson enables insight into agent reasoning without modifying the primary agent or underlying model. However, the debugging agents used in this study are illustrative rather than optimized, designed to validate trace utility rather than maximize performance. Moreover, \watson is an observability layer, not a reasoning optimizer, and its effectiveness ultimately depends on the interpretability and completeness of the extracted traces.

\begin{revision}
\section{Applications, Limitations, and Threats to Validity}
\label{sec:threats}
In this section, we discuss potential applications of \watson in practical development settings and examine its limitations and threats to validity.

\subsection{Applications in Practical Settings}
\label{sec:threats:applications}
A promising avenue for the adoption of \watson is its integration into observability platforms designed for agentic workflows. In such settings, workflow executions are typically instrumented to capture detailed traces of agent behavior, including inputs, outputs, and intermediate decision steps. \watson can be incorporated into these platforms as an on-demand reasoning reconstruction service, enabling users to request surrogate reasoning for specific agent calls when deeper insight is required. This capability extends traditional observability by augmenting traces with interpretable reasoning paths, thereby facilitating debugging, auditing, and comprehension of complex multi-agent workflows. Crucially, \watson achieves this without modifying or delaying the execution of the original agent, making it suitable for deployment in developer environments (e.g., IDEs), continuous integration and deployment pipelines, or production-grade observability dashboards. Such integration illustrates \watson{}’s potential to complement existing software engineering practices with cognitive observability, thereby improving transparency and trust in LLM-powered agents.

\subsection{Limitations}
\label{sec:threats:limitations}
While the integration scenarios highlight \watson’s potential value, it is important to consider its performance trade-offs. \watson adds no runtime overhead to the primary agent, since reasoning reconstruction is applied post hoc to captured traces. However, generating multiple completions with the surrogate agent to recover diverse reasoning paths and produce a meta-reasoning summary (Section~\ref{sec:watson:overview}) incurs computational cost. In our experiments, we generated 10 reasonings per instance to balance coverage and expense. \watson{}’s computational cost then scales with the number of reasoning traces generated and the computation of prompt attribution scores used for trace verification, which scale with prompt length and component decomposition. Both \fim and \repcot discard traces misaligned with these scores, regenerating until the desired number of valid traces is reached, further contributing to overhead. 

Importantly, the number of surrogate completions provides a direct trade-off mechanism. Fewer completions reduce cost and latency but may miss important reasoning paths, while more completions increase coverage and confidence at the expense of computational resources. However, these computations do not affect the runtime of the observed agent, as they can be performed in parallel or post hoc. The exact cost depends on prompt complexity, the number of surrogate completions, and prompt segmentation for attribution scoring, allowing \watson to adapt to different deployment contexts, from lightweight development workflows to resource-intensive audits. Despite this flexibility, \watson cannot guarantee complete coverage of all plausible reasoning paths. Even with higher sampling, some trajectories may remain unexplored, especially for complex tasks with large search spaces. Consequently, the reconstructed reasoning should be interpreted as a representative sample rather than an exhaustive account of the agent’s internal processes.

A further limitation of \watson arises from its reliance on an \llm-as-a-judge for reasoning validation (Section~\ref{sec:watson:verifyingreasoning}). After \promptexp identifies the most and least influential prompt components, the judge checks that each generated reasoning trace aligns with high-attribution components without overemphasizing low-attribution ones. While this grounds validation in the agent’s decision factors, the judge can misclassify traces, either rejecting plausible reasoning or accepting partially misaligned ones. We mitigate this by using a strong judge model and combining attribution scores with explicit rules (Section~\ref{sec:watson:verifyingreasoning}), but some subjectivity and bias remain. The fidelity of verified traces thus depends on both surrogate coverage and judge evaluation, highlighting directions for improved calibration, robustness, and reliability in future work.

Finally, it is important to note that \watson has so far only been evaluated on text-based reasoning agents, and its applicability to other types of agents remains uncertain. Extending \watson to these domains would likely require additional mechanisms for capturing and interpreting non-textual context, as well as adapting the verification procedures to handle richer forms of agent reasoning. Consequently, the framework’s generalizability beyond text-based reasoning agents should be considered a limitation and an area for future research.

\subsection{Threats to Validity}
\label{sec:threats:validity}
Several factors may influence the reliability and generalizability of the experimental results reported for \watson. First, the fidelity of the surrogate agent is critical: if the surrogate does not accurately mirror the primary agent’s reasoning behavior, the reconstructed traces may not reflect the true internal processes, limiting the validity of our conclusions. Second, the evaluation depends on the \llm-as-a-judge used for reasoning verification. Biases or misclassifications by this model can affect which reasoning traces are retained, potentially influencing the resulting meta-reasoning summaries. Third, the framework has only been evaluated on text-based reasoning agents, and it remains uncertain how well the results would generalize to multimodal, embodied, or highly interactive agents. Finally, reproducibility is a potential concern: variations in surrogate configurations, the number of completions generated, or the choice of judge model may yield different reasoning reconstructions, affecting consistency across experiments. By explicitly acknowledging these threats, we aim to provide a transparent assessment of the contexts in which \watson{}’s results can be confidently interpreted.

\end{revision}

\section{Conclusion}
\label{sec:conclusion}
In this paper, we introduced the concept of \emph{cognitive observability} as a crucial advancement in the observability of agentic software powered by foundation models. Traditional observability techniques fall short in these systems due to the opaque, non-deterministic reasoning processes of \llms. To address this challenge, we proposed and implemented \watson, a novel framework that retrospectively recovers implicit reasoning traces with high fidelity while preserving the efficiency of standard \llms.

Our evaluation demonstrates that \watson enables both manual debugging and automated runtime correction, effectively exposing the ``why'' behind agent decisions in a manner that was previously inaccessible without affecting the agent's behavior or outputs. Through empirical validation on real-world agent tasks, including \del{\mmlu and \acr}\rev{\mmlu, and \acr and \oh} on the \swebenchlite benchmarks, \watson facilitates improved reasoning transparency and supports enhanced reliability and controllability of \agentware. Overall, \watson offers a practical step toward addressing the observability challenges of \llm-powered agents, helping developers better interpret and improve agent behavior. Future work should explore broader generalization across agent types and tasks, more sophisticated use of reasoning traces for intervention or retraining, and real-time observability in production environments. As \llms continue to take on complex roles in \agentware, advancing tools for cognitive observability will be critical to supporting their safe and effective deployment.

\printbibliography

\end{document}